\documentclass[10pt,twocolumn,letterpaper]{article}

\usepackage[pagenumbers]{cvpr} 

\usepackage{listings}
\usepackage{multirow}
\usepackage{caption}

\definecolor{cvprblue}{rgb}{0.21,0.49,0.74}
\usepackage[pagebackref,breaklinks,colorlinks,allcolors=cvprblue]{hyperref}

\def\paperID{22268} 
\def\confName{CVPR}
\def\confYear{2026}

\title{PPTBench: Towards Holistic Evaluation of Large Language Models for PowerPoint Layout and Design Understanding}

\author{
Zheng Huang$^{1,2}$ \quad
Xukai Liu$^{1,2}$ \quad
Tianyu Hu$^{1}$ \quad
Kai Zhang$^{1,2}$ \quad
Ye Liu$^{1,2}$ \\
$^{1}$ University of Science and Technology of China\\
$^{2}$ State Key Laboratory of Cognitive Intelligence\\
{\tt\small \{huangzheng, chthollylxk, tianyuhu\}@mail.ustc.edu.cn} \\
{\tt\small kkzhang08@ustc.edu.cn \quad yeliu.liuyeah@gmail.com}
}

\begin{document}
\maketitle
\begin{abstract}
PowerPoint presentations combine rich textual content with structured visual layouts, making them a natural testbed for evaluating the multimodal reasoning and layout understanding abilities of modern MLLMs.
However, existing benchmarks focus solely on narrow subtasks while overlooking layout-centric challenges, which are central to real-world slide creation and editing.
To bridge this gap, we introduce PPTBench, a comprehensive multimodal benchmark for 
evaluating LLMs on PowerPoint-related tasks. Leveraging a diverse source of 958 PPTX 
files, PPTBench evaluates models across four categories with 4,439 samples, including Detection, Understanding, Modification, and Generation. 
Our experiments reveal a substantial gap between semantic understanding and visual–layout reasoning in current MLLMs: models can interpret slide content but fail to produce coherent spatial arrangements.
Ablation and further analysis show that current MLLMs struggle to combine visual cues with JSON-based layout structures and fail to integrate visual information into their API planning ability.
And case studies visually expose systematic layout errors such as misalignment and element overlap.
These findings provides a new perspective on evaluating VLLMs in PPT scenarios, highlighting challenges and directions for future research on visual–structural reasoning and coherent slide generation.
All datasets and code are fully released to support reproducibility and future research:
\url{https://github.com/Gastronomicluna/PPTBench-Eval}.
\end{abstract}


\section{Introduction}
\label{sec:intro}

Recent advances in MLLMs have enabled AI systems to jointly process and reason over text and vision, demonstrating strong performance in various tasks such as visual question answering, document and chart understanding, and code-based tool use~\cite{koh2024visualwebarena,yang2024swe,szot2025multimodal}.
To systematically evaluate the capabilities of MLLMs, a series of benchmarks such as MMMU~\cite{MMMU}, 
MMMU-Pro~\cite{MMMU-Pro}, MathVista~\cite{MathVista}, AI2D~\cite{hiippala2021ai2d}, and MMStar~\cite{chen2024rightwayevaluatinglarge} have been introduced, collectively driving rapid progress in vision–language research.
Among various multimodal application domains, PowerPoint presentations represent a unique and practical domain for multimodal reasoning. 
They contain rich textual content organized within highly structured visual layouts, where spatial relationships strongly influence how information is interpreted.
Recent commercial systems, such as Canva\footnote{https://www.canva.com/create/ai-presentations/}, 
Microsoft Copilot\footnote{https://support.microsoft.com/en-us/copilot-powerpoint}, 
and Beautiful.AI\footnote{https://www.beautiful.ai}, 
already employ VLLM for automating slide generation, 
highlighting the real-world demand for intelligent PPT assistants.

In contrast, academic research on PowerPoint reasoning and automation remains limited. 
Existing researches such as PPTC~\cite{guo-etal-2024-pptc}, SLIDESBENCH~\cite{ge2025autopresent}, and PPTAgent~\cite{zheng2025pptagent} have introduced benchmarks focusing on individual sub-tasks like API-based modification or natural language–to–slide generation. 
However, existing datasets suffer from two fundamental limitations.
First, they address only narrow and isolated subtasks, overlooking the breadth of operations involved in real-world presentation creation, editing, and refinement. This prevents a holistic assessment of model capabilities.
Second, current benchmarks largely ignore the crucial visual aspects of presentations. They provide limited support for layout reasoning, spatial manipulation, and design-level understanding, which that are central to producing coherent and professionally structured slides.

\begin{figure*}[h]
	\centering
	\includegraphics[width=0.90\textwidth]{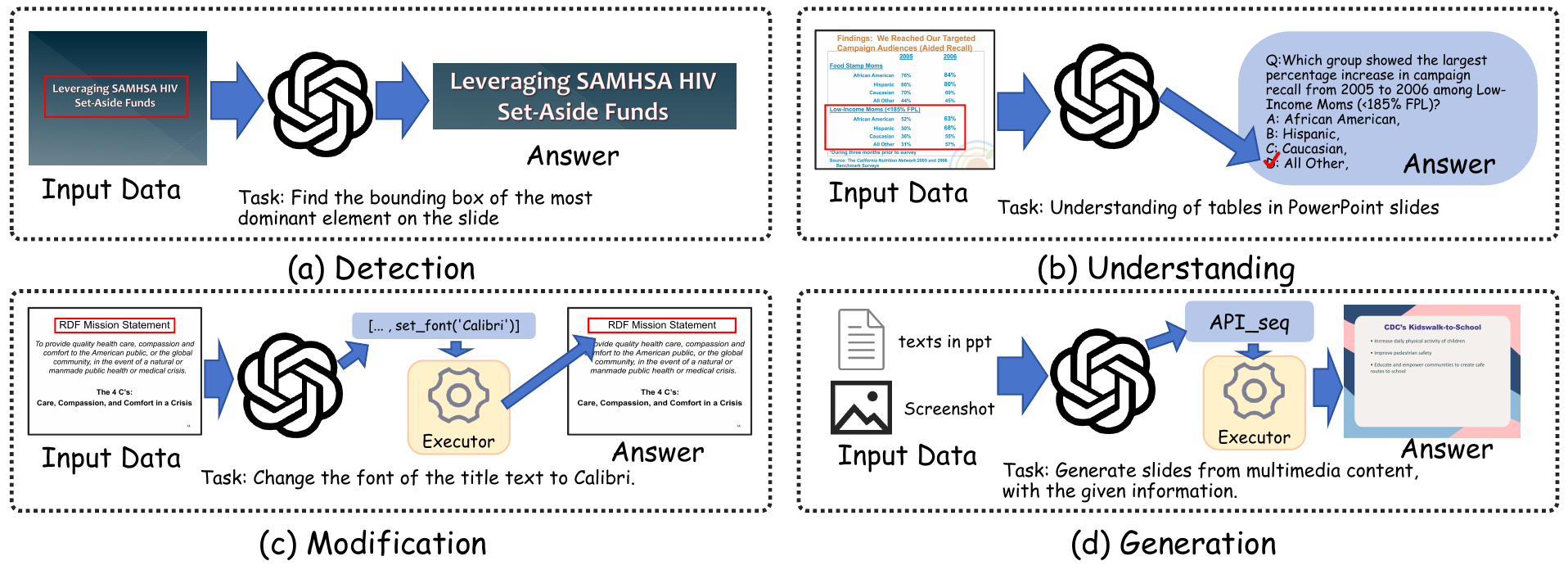}
	\caption{Task examples of the four categories in PPTBench on PowerPoint presentations.}\label{fig:pptbench-example}
\end{figure*}

To fill this evaluation gap, we present PPTBench, a comprehensive benchmark specifically designed to assess the capabilities of VLLMs on PowerPoint-related tasks. Drawing from common human-presentation interactions, we organize these tasks into four progressive categories with 4,439 samples, as illustrated in Figure~\ref{fig:pptbench-example}:
(a) \textbf{Detection}: Locating and extracting specific elements or information within presentations.
(b) \textbf{Understanding}: Inferring implicit relationships, semantic hierarchies, and structural organization within slides beyond surface-level element detection. 
(c) \textbf{Modification}: Implementing precise edits to content and layout based on specific requirements.
(d) \textbf{Generation}: Creating coherent, well-structured slides or complete presentations from scratch.
Furthermore, PPTBench explicitly evaluates visual and layout reasoning through tasks that require interpreting spatial relationships, element hierarchy, and slide structure.
To ensure real world applicability, we curated over 950 professional PowerPoint presentations spanning diverse domains and styles, forming a robust basis for evaluating LLM performance in practical scenarios.
To enable models to simultaneously access both visual information and structured numerical layout data, 
each presentation is processed into a dual-modality input format—structured JSON files encoding textual and layout metadata, paired with corresponding slide screenshots that provide visual grounding. 
Overall, these features make PPTBench a comprehensive and realistic benchmark for evaluating VLLMs' capabilities in multimodal reasoning, structural understanding, and end-to-end slide manipulation within PowerPoint.



Our experiments find a clear gap between semantic understanding and visual reasoning across a wide range of models.
Models can interpret slide content but fail to translate the understanding into coherent spatial layouts.
Ablation results further show that current MLLMs struggle to fuse visual information with JSON-based layout structures in PPT tasks.
Meanwhile, additional experiments demonstrate that template-guided generation significantly improves structural coherence.
But this also suggests that current MLLMs lack the ability to integrate sequential action planning with visual layout understanding, which is essential for executing API operations correctly and producing well-organized slides.
Case studies provide a direct illustration of this limitation, showing that models frequently overlook layout constraints, misplace elements, and cause content overlaps during execution.
These findings provide a new perspective on evaluating VLLMs in PowerPoint scenarios, highlighting challenges and directions for future research on visual–structural reasoning and coherent slide generation.

\section{Related Work}
\label{sec:related work}

\subsection{Evaluation of LLMs in the PowerPoint}

Large-scale MLLMs have achieved remarkable progress across a variety of multimodal tasks. To systematically assess their capabilities, numerous benchmarks have been introduced in recent years. General-purpose benchmarks such as LiveBench~\cite{livebench} evaluate reasoning, coding, and instruction-following abilities, while multimodal-focused datasets like MMMU~\cite{MMMU} and MMMU-Pro~\cite{MMMU-Pro} extend evaluation to visual understanding. In addition, several domain-specific datasets, such as ChartQA~\cite{masry-etal-2022-chartqa} for chart comprehension and APIGen~\cite{liu2024apigenautomatedpipelinegenerating} for multimodal API interactions, serve as valuable resources for model training and fine-grained capability assessment.

Recently, several benchmarks have explored PowerPoint multimodal tasks. PPTC~\cite{guo-etal-2024-pptc} evaluates LLMs’ ability to modify slides through API-based operations. PPTC-R~\cite{zhang2024pptc} focuses on robustness against adversarial instructions and software variations. 
SLIDESBENCH~\cite{ge2025autopresent} defines the NL-to-Slide task for generating presentation slides from natural language. 
PPTAgent~\cite{zheng2025pptagent} formulates document-to-presentation generation via structured, edit-based workflows with layout reasoning.
However, existing multimodal benchmarks exhibit two critical limitations: 
(1) Most prior work isolates specific subtasks, without providing a unified evaluation across understanding, reasoning, and design. 
(2) Many benchmarks lack visual grounding, as they do not incorporate visual context or layout information. 
To address these limitations, we introduce PPTBench, a comprehensive benchmark that unifies multimodal reasoning and layout-aware generation for PowerPoint-related tasks.

\subsection{Vision-Language Large Models}

Recent advances in large language models (LLMs)~\cite{achiam2023gpt,dubey2024llama,touvron2023llama} and vision–language models (VLMs)~\cite{alayrac2022flamingo,yang2024posterllava} have greatly accelerated progress in multimodal understanding and reasoning.
The emergence of multimodal large language models (MLLMs), including closed-source models such as GPT-4o~\cite{achiam2023gpt} and Gemini~\cite{team2023gemini}, as well as open-source models like LLaVA~\cite{liu2023visual}, MiniCPM-V~\cite{hu2024minicpm}, InternLM-XComposer~\cite{zhang2023internlm}, Qwen-VL~\cite{wang2024qwen2}, and MiniGPT-4~\cite{zhu2023minigpt}, has shown impressive multimodal understanding and reasoning capabilities, further highlighting the potential of scaling unified vision–language pretraining.

Agents based on LLMs and VLMs have emerged as powerful paradigms and have been widely adopted across diverse multimodal domains. 
For example, web navigation tasks~\cite{koh2024visualwebarena,yao2022webshop,zhou2023webarena} investigate structured visual reasoning and grounded interaction on webpages; 
software engineering agents~\cite{yang2024swe,yang2024swem} focus on code understanding and automated software manipulation; 
Embodied AI systems~\cite{szot2025multimodal,cheng2025embodiedeval} evaluate how MLLMs plan and act through vision-guided control.
Despite notable progress in multimodal reasoning, existing MLLMs still lack the capability to understand the structured layout and compositional organization of PowerPoint slides, which are essential for accurate content interpretation and manipulation.
This gap motivates our development of \textbf{PPTBench}, which systematically evaluates MLLMs on PowerPoint-related multimodal reasoning and manipulation.

\begin{figure}[t]
	\centering
	\includegraphics[width=0.95\columnwidth]{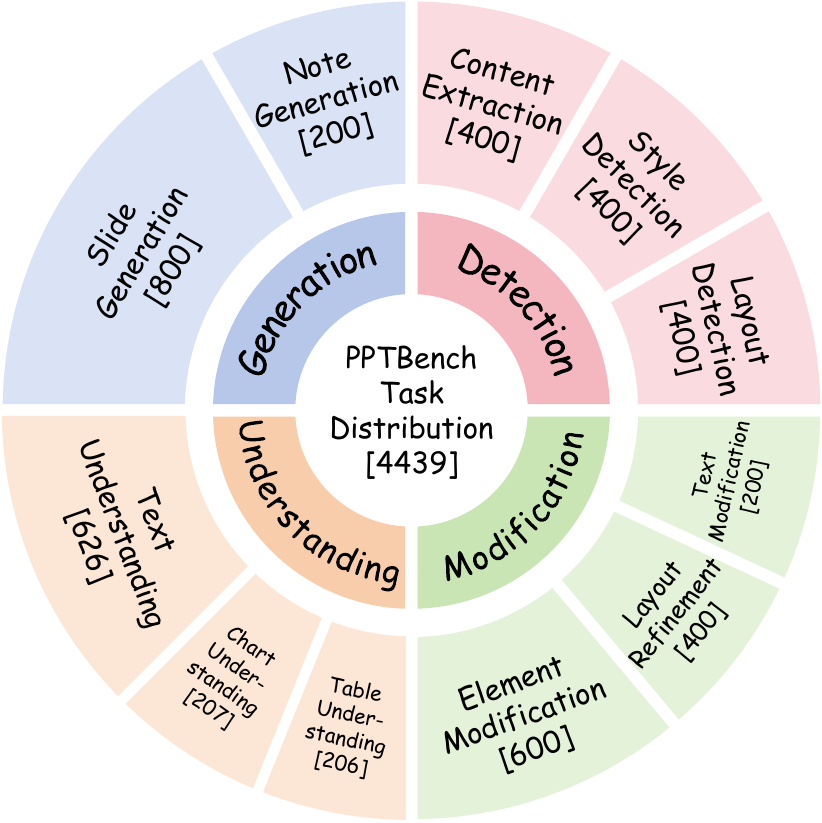}
	\caption{Distribution of task categories in PPTBench, showing the number of samples under each sub-task.}\label{fig:pptbench-distribution}
\end{figure}

\section{PPTBench}

\subsection{Overview}

In this section, we present PPTBench, a comprehensive benchmark designed to evaluate MLLMs across PowerPoint-related tasks. The benchmark is organized into four major categories: detection, understanding, modification, and generation. As illustrated in Figure~\ref{fig:pptbench-distribution}, 
PPTBench contains 4,439 samples spanning 11 sub-tasks, each targeting a distinct aspect. More detailed statistics are provided in Appendix~\ref{App: datasets}.
We next describe each task category and its corresponding construction in detail.

\subsection{Benchmark Designs}

\subsubsection{Detection Tasks}

Detection tasks evaluate the fundamental abilities of LLMs in perceiving slide elements, focusing on extracting content, detecting style, and analyzing layout, which are prerequisites for more complex operations.

\textbf{(1) Content Extraction:} Identifying and extracting structured textual information from 
slides, such as distinguishing between titles, subtitles, bullet points, paragraphs, and 
footnotes. 
This capability provides the foundation for PPT understanding and layout analysis.

\textbf{(2) Style Detection:} Recognizing and categorizing formatting attributes that convey visual 
emphasis and information hierarchy, including font properties (typeface, weight, size, color), 
text effects (highlighting, shadowing), and stylistic variations (italics, bold, 
underline). This capability is essential for maintaining design consistency 
and understanding intended emphasis in presentations.

\textbf{(3) Layout Detection:} Analyzing the spatial layout and geometric relationships among 
slide elements.  
This includes detecting issues such as overlapping shapes, elements extending beyond slide boundaries, misaligned components, and inconsistent spacing, as well as locating the bounding boxes of dominant elements or identifying the highest object on the slide.
This capability is essential for keeping slides visually organized.

\subsubsection{Understanding Tasks}

Understanding tasks evaluate whether LLMs can achieve a deeper comprehension of slide content beyond surface recognition, enabling them to infer underlying conclusions within a slide. 
This involves interpreting textual content, reasoning through charts and tables to achieve a holistic understanding of slide semantics.

\textbf{(1) Text Understanding:} Inferring implicit textual information in slides. 
This involves identifying main topics, summarizing textual information, and recognizing hierarchical relationships among text elements.
This ability is crucial for understanding the core content of the slides.

\begin{table}[t!]
\centering
\caption{Usage of Elements Across Different Task.}
\label{tab:element_usage}
\resizebox{\linewidth}{!}{
\begin{tabular}{lcccc}
\toprule
\textbf{Task} & \textbf{S (Slide Data)} & \textbf{Q (Task Query)} & \textbf{A (API List)} & \textbf{L (Label)} \\ 
\midrule
Detection      & $\checkmark$ & $\checkmark$ & $\times$ & $\checkmark$ \\
Understanding  & $\checkmark$ & $\checkmark$ & $\times$ & $\checkmark$ \\
Modification   & $\checkmark$ & $\checkmark$ & $\checkmark$ & $\checkmark$ \\
Generation     & $\times$ & $\checkmark$ & $\checkmark$ & $\times$ \\
\bottomrule
\end{tabular}}
\vspace{0mm}
$\checkmark$ denotes the element is used; 
$\times$ denotes it is not used.
\end{table}

\textbf{(2) Chart Understanding:} Interpreting and reasoning over the graphical chart within slides.
This involves identifying chart types, analyzing data series, and interpreting axes and legends to recognize patterns, correlations, and anomalies across diverse visualization formats.
This capability is essential for supporting semantic reasoning over visualized information in presentations.

\textbf{(3) Table Understanding:} Reasoning over structured tabular data within slides. 
This involves analyzing table structure, interpreting headers and cell contents, and recognizing logical relationships among rows, columns, and categories to identify hierarchical organization within structured data. 
This capability is essential for accurately interpreting detailed information and supporting reliable quantitative analysis in slides with tables.

\subsubsection{Modification Tasks}
Modification tasks evaluate the capability of LLMs to modify existing slides based on user instructions, including editing individual elements, updating textual content, and refining overall layout structures.

\textbf{(1) Element Modification:} Performing fundamental object-level actions such as adding, deleting, resizing, or repositioning shapes.
This involves the understanding of user instructions and API calls, serving as a foundational capability for achieving higher-level content modification.

\textbf{(2) Text Modification:} Editing textual elements based on instructions,
such as correcting grammar, refining tone, or adapting phrasing to new contexts.
This capability reflects the model’s ability in contextual understanding and semantic preservation in PPT presentation scenarios.
  
\textbf{(3) Layout Refinement:} Adjusting the spatial organization of elements within slides. 
This encompasses operations such as automatic alignment, resolving out-of-boundary element placement, and optimizing inter-element spacing.
This capability ensures that slides remain clear, well-organized, and visually consistent.

\subsubsection{Generation Tasks}
Generation tasks assess the capability of LLMs to create new slides from scratch, guided by specific requirements or source materials. 
These tasks require the model to synthesize information, structure content coherently, and design visually appropriate layouts.

\textbf{(1) Note Generation:} 
Generating explanatory notes for each slide, capturing the key ideas and intended message of the presentation.
This involves integrating information across multiple slides, summarizing key ideas, and generating coherent speaker notes that reflect the overall logical flow of the presentation.
This capability demonstrates the model’s global understanding and its ability to produce contextually connected, audience-oriented explanations.

\textbf{(2) Slide Generation:} Constructing a complete slide from diverse input modalities such as screenshots or text. 
This involves inferring and reproducing spatial layouts from multimodal inputs while applying appropriate formatting, styling, and design conventions. 
As the most comprehensive task in PPTBench, this capability integrates content comprehension, layout organization, and visual design to produce clear and professional slides.

\subsection{Data Construction}
PPTBench utilizes 958 PowerPoint presentations sourced from government archives. Each instance is represented as a tuple $(S,Q,A,L)$ to ensure consistency, as shown in Table~\ref{tab:element_usage}. We next detail the construction of each component.

\subsubsection{Slide Data $S$}
$S$ represents the source slide data. 
However, Raw PowerPoint files are unsuitable for direct use with LLMs.
Despite adopting the Open XML\footnote{https://support.microsoft.com/en-us/office/open-xml-formats-and-file-name-extensions-5200d93c-3449-4380-8e11-31ef14555b18}
 standard, they contain excessive metadata and represent layouts only through coordinate values, which hinders direct semantic reasoning.
Therefore, structured extraction and normalization are required.

To overcome these challenges, we construct a dual-input representation:
\textbf{Structured JSON} encodes shape types, textual content, and spatial attributes (position and size);  
\textbf{Slide Screenshot} provides direct visual grounding. 
This representation preserves both structural and visual information while discarding unnecessary XML complexity, enabling LLMs to process PowerPoint slides efficiently.

\subsubsection{Task Query $Q$}
$Q$ represents the task-specific instruction that guides the model to perform a particular operation, including general query and multiple-choice question.

\paragraph{General Query.}
The general query contains only the user instruction.
To ensure these instructions are both task-relevant and executable, we construct a small set of natural-language templates and extraction algorithms for each major task category. Variable placeholders in templates are then automatically extracted.
General queries are applied to the Detection, Modification, and Generation tasks.

\paragraph{Multiple-Choice Question.}
For the \textit{Understanding} tasks, we adopt the multiple-choice question since the correct answers are often implicit, making direct evaluation unreliable.
Specifically, we use LLMs generate multiple-choice questions based on the slides, and then human reviewers verify both the questions and answers. Two independent annotators were involved in this process, achieving a 98\% agreement rate to ensure annotation quality.


\subsubsection{API List $A$}
To ensure controllable and executable manipulation of PowerPoint slides, PPTBench employs an API-based interaction paradigm that translates natural-language instructions into precise, verifiable editing operations.
Following PPTC~\cite{guo-etal-2024-pptc}, we adopt a standardized API schema comprising 17 atomic PowerPoint functions shown in Appendix~\ref{App:Templates and API}, each with explicit parameter definitions specifying shape identifiers, coordinates, colors, and style attributes.
During inference, the model generates a structured sequence of API calls ($A_{\text{seq}}$), which are executed by a deterministic slide executor to produce the modified or newly generated slides.

\subsubsection{Label $L$}
$L$ denotes the ground-truth for each task.
PPTBench derives labels by reverse-engineering each task from the JSON representation, ensuring structural accuracy, semantic consistency, and reproducibility.

All PowerPoint files are first converted into compact JSON structures that preserve the textual, geometric, and stylistic attributes of every element.
Labels for downstream tasks are then generated directly from these JSON files, guaranteeing alignment with the true visual and semantic content of the slides. Specifically, 
\textbf{For Detection:} Labels are directly read from JSON fields corresponding to element type, position, and format. 
\textbf{For Understanding:} Labels are obtained through a semi-automatic process, visual parsing from JSON, LLM-based question generation, and human verification, to ensure factual correctness. 
\textbf{For Modification:} Labels correspond to post-editing slide states, produced by executing controlled API operations on the JSON structure for deterministic and verifiable outcomes.
For the Generation tasks, the source PPT can serve as a reference, but it is not used as a label due to the inherently subjective of generation tasks.


To ensure label reliability, all generated labels undergo automated consistency checking and partial human auditing, ensuring that $L$ serves as a faithful and reproducible ground truth across tasks. 

In Appendix~\ref{App:data construction}, we provide additional details of the data construction to better understand the overall pipeline.


\begin{table*}
    \centering
    \caption{\label{tab:model-performance}Comparison of Model Performance across Task in PPTBench.}
    \begin{tabular}{l@{\hspace{8pt}}c@{\hspace{8pt}}c@{\hspace{8pt}}c@{\hspace{8pt}}c@{\hspace{8pt}}c}
        \toprule
        \textbf{Model} &
        \textbf{Overall} &
        \textbf{Detection} &
        \textbf{Understanding} &
        \textbf{Modification} &
        \textbf{Generation} \\
        \midrule
        \multicolumn{6}{c}{Closed-source Models} \\
        \midrule
        OpenAI GPT-4o & \textbf{76.79} & \textbf{95.92} & 94.03 & \textbf{86.95} & 30.25 \\
        Gemini-2.0-Flash & 72.16 & 82.25 & \textbf{95.28} & 78.27 & \textbf{32.82} \\
        \midrule
        \multicolumn{6}{c}{Open-source Models} \\
        \midrule
        Llama3.2-Vision (11B) & 41.84 & 66.75 & 84.79 & 11.70 & 4.11 \\
        Llama3.2-Vision (90B) & 63.22 & 84.58 & 89.51 & 51.19 & 27.59 \\
        LLaVA-1.5 (13B) & 31.92 & 56.58 & 70.93 & 0.17 & 0.00 \\ 
        LLaVA-1.5 (34B) & 38.85 & 65.61 & 79.50 & 10.27 & 0.00 \\ 
        MiniCPM-V (8B) & 32.69 & 45.75 & 83.06 & 1.94 & 0.00 \\
        Gemma3 (12B) & 50.49 & 79.00 & 75.46 & 35.73 & 11.75 \\
        \bottomrule
    \end{tabular}
\end{table*}


\section{Experiments}

\subsection{Experimental setup}
We evaluate several state-of-the-art LLMs from both closed-source and open-source models, covering a range of model sizes and architectures.
For closed-source models,
our evaluation includes:
GPT-4o~\cite{gpt4o} from OpenAI and Gemini-2.0-Flash~\cite{gemini2-google} from Google.
For open-source models, we include:
Llama3.2-Vision (11B, 90B)~\cite{grattafiori2024llama3herdmodels, llama32} from Meta,
LLaVA-1.5 (13B, 34B)~\cite{liu2023visual, Liu_2024_CVPR}, 
MiniCPM-V (8B)~\cite{hu2024minicpm, yao2024minicpmvgpt4vlevelmllm} and 
Gemma-3 (12B)~\cite{team2025gemma}
To ensure reproducibility, we set the temperature to 0.

\subsection{Metrics}
\label{sec:metrics}

To ensure objective and consistent evaluation, we design specific metrics for each task category as follows:

\paragraph{Detection}
Let M be the model, 
given $(S, Q)$, the model produces an output $A = M(S, Q)$, which is compared with the ground-truth label $L$ by exact matching:
\begin{equation}
\text{Acc}_{\text{det}} = \frac{1}{N} \sum_{i=1}^{N} \mathbb{I}(A_i = L_i)
\end{equation}
where $\mathbb{I}(\cdot)$ is the indicator function and $N$ denotes the total number of detection samples.

\paragraph{Understanding}
The Task query $Q$ includes four candidate answers $\mathcal{E} = \{\text{opt}_1, \text{opt}_2, \text{opt}3, \text{opt}4\}$. 
The model selects one option $A = M(S, Q)$. 
Accuracy is computed based on the correct classification of the true label $L \in \mathcal{E}$:
\begin{equation}
\text{Acc}_{\text{und}} = \frac{1}{N}\sum_{i=1}^{N} \mathbb{I}(A_i = L_i).
\end{equation}

\paragraph{Modification}
Model $M$ generates a sequence of API calls to modify specific elements $A_{\text{seq}} = M(S, Q, A)$. Let $E(\cdot, \cdot)$ denote an executor function that applies an API sequence to a slide. 
After execution, the resulting slide is represented by its structured JSON file $S^{*} = E(A_{\text{seq}}, S)$. 
Modification performance is evaluated by exact match between the resulting JSON structure and the ground-truth label $L$:
\begin{equation}
\text{Acc}_{\text{mod}} = \frac{1}{N} \sum_{i=1}^{N} \mathbb{I}(S^{*}_i = L_i).
\end{equation}
This metric ensures that every structural attribute (element type, position, color, and style) in the modified slide exactly matches the reference output.

\paragraph{Generation}
The model $M$ is required to create entirely new slides based on the input $S'$. 
The model’s objective is to generate a sequence of API calls 
$A_{\text{seq}} = M(S', Q, A)$, which, when executed, produces a new slide $S^{*}$ that effectively conveys the information contained in $S'$.

Due to the inherently subjective and multi-faceted nature of slide design quality, we adopt a \textit{LLM-as-a-Judge}~\cite{ge2025autopresent} evaluation paradigm. Using \textit{Gemini-2.0} as the evaluation model, each generated slide is rated according to a standardized six-point rubric:

\begin{itemize}
    \item[0] \textbf{(Very Poor):} No meaningful content or completely unreadable/confusing.
    \item[1] \textbf{(Poor):} Message unclear, poor design, severe issues with layout or content.
    \item[2] \textbf{(Fair):} Basic message present but lacking clarity, coherence, or visual polish.
    \item[3] \textbf{(Good):} Message clear, design acceptable, minor improvements needed.
    \item[4] \textbf{(Very Good):} Strong clarity and design, effective visual communication.
    \item[5] \textbf{(Excellent):} Highly professional, well-balanced, visually engaging and impactful.
\end{itemize}

Each score $\text{Score}_i \in \{0,1,2,3,4,5\}$ is mapped to a percentage scale for interpretability:
\begin{equation}
\text{Score}_{\text{gen}} = \frac{1}{N}\sum_{i=1}^{N} \text{Score}_i \times 20
\end{equation}
where $N$ denotes the total number of generation samples. 
For more metrics details, please refer to the Appendix~\ref{app:generation}.

\subsection{Main Findings}

Table~\ref{tab:model-performance} presents the comprehensive evaluation results across all models and tasks. 
We observed the following interesting phenomenon:

\textbf{(1) Overall performance.} 
Closed-source models consistently outperform open-source counterparts across all evaluated task categories. 
Specifically, GPT-4o achieves the highest scores in \textit{Detection} (95.92) and \textit{Modification} (86.95), while Gemini-2.0-Flash leads in \textit{Understanding} (95.28). 
Nevertheless, both models exhibit notable weaknesses in the \textit{Generation} task, where Gemini-2.0-Flash reaches only 32.82 and GPT-4o reaches only 30.25. 

By contrast, open-source models also demonstrate reasonably strong performance in \textit{Detection} and \textit{Understanding}, showing that they retain fundamental capabilities in content extraction and semantic comprehension.
However, they fail in \textit{Modification} and \textit{Generation}, where the ability of precise API planning is required. This gap highlights their limited abilities in multi-step planning required for complex slide editing and generation.

Performance also scales positively with model size across all evaluated model families.
For example, in Llama3.2-Vision family, the 90B variant substantially surpasses the 11B version across all tasks, 
most notably in Modification (51.19\% vs. 11.70\%). 
This trend suggests that larger models possess stronger reasoning abilities, enabling them to handle more complex PPT-related tasks.

\textbf{(2) A significant gap remains between "semantic comprehension" and "structural manipulation."}
As shown in Table~\ref{tab:model-performance}, the four tasks form a clear difficulty hierarchy:
\textit{Detection} $<$ \textit{Understanding} $<$ \textit{Modification} $<$ \textit{Generation}.
While most models perform well on the first two categories, performance drops sharply on Modification tasks, which require translating user intent into precise, executable API operations.
This difficulty gap is especially evident in smaller models such as MiniCPM-V (8B) and LLaVA-1.5 (13B), whose scores nearly collapse to zero.
Generation poses an even greater challenge, as models consistently fail to produce coherent, well-structured slides from raw inputs. Even the best-performing model achieves only around 33\%, while several open-source models score near zero, underscoring the difficulty of this task.
Overall, these results reveal that current MLLMs struggle to combine visual understanding with high-level multimodal planning in PPT-related tasks.



\begin{figure*}[h]
	\centering
	\includegraphics[width=0.90\textwidth]{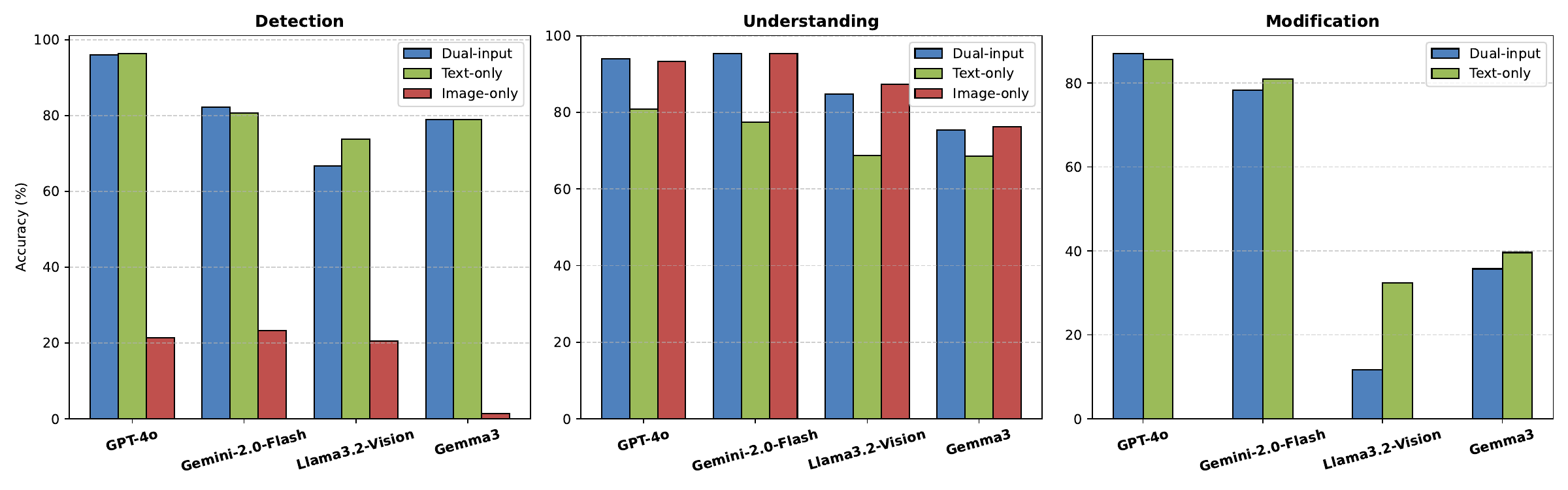}
	\caption{Ablation study on selected tasks. Note that Image-only is not applicable for Modification, and Generation need no input source.}
    \label{fig:ablation-study}
\end{figure*}

\begin{table*}[htbp]
\centering
\caption{Effect of CoT across task categories.}
\label{tab:ablation-cot}
\begin{tabular}{lcccccccc}
\hline
\multirow{2}{*}{\textbf{Model}} 
& \multicolumn{2}{c}{\textbf{Detection}} 
& \multicolumn{2}{c}{\textbf{Understanding}} 
& \multicolumn{2}{c}{\textbf{Modification}} 
& \multicolumn{2}{c}{\textbf{Generation}} \\ 
\cline{2-9}
& w/o CoT & w/ CoT
& w/o CoT & w/ CoT
& w/o CoT & w/ CoT
& w/o CoT & w/ CoT \\ 
\hline
OpenAI GPT-4o         & 95.92 & 95.08  & 94.03 & 94.51 & 86.95 & 84.01 & 30.25 & 39.36 \\
Gemini-2.0-Flash      & 82.25 & 94.92 & 95.28 & 94.71 & 78.27 & 77.27 & 32.82 & 37.91 \\
LLaVA1.5 (13B)        & 56.58 & 38.42 & 70.93 & 63.52 & 0.17 & 9.97 & – & – \\
Gemma3 (12B)          & 79.00 & 78.92 & 75.46 & 58.02 & 35.73 & 57.64 & 11.75 & 33.15 \\
\hline
\end{tabular}%
\end{table*}

\subsection{Ablation Study}

To further examine the contribution of different input modalities in PPTBench, 
we conduct ablation studies focusing on the effects of structured JSON inputs and visual slide screenshots. 
Note that Modification tasks require editing the original JSON representation of slides, so the Image-only configuration is not applicable. 
Similarly, Generation tasks start from blank slides and thus are excluded from ablation study.

As shown in Figure~\ref{fig:ablation-study}, the dual-input configuration generally achieves the best overall performance across most models and tasks. 
Specifically, Detection and Modification tasks benefit substantially from JSON-based structural metadata, 
as they rely on precise positional encoding and shape-level attributes in the JSON data.  
In contrast, Understanding tasks show smaller gains from JSON input but exhibit stronger sensitivity to visual information of slide screenshots, as this category includes chart and table understanding, both of which rely heavily on visual cues and spatial relationships.
This indicates that MLLMs leverage holistic visual semantics for comprehension while depending on explicit structure for accurate manipulation.

Interestingly, we also observe several instances where removing one modality unexpectedly leads to better performance.  
This phenomenon is more pronounced in smaller models in Modification tasks, suggesting that limited multimodal fusion capability may hinder their ability to align spatial and semantic representations.  



\subsection{Discussion}

\subsubsection{Impact of Chain-of-Thought}
\label{exp:cot}
To examine the impact of Chain-of-Thought (CoT) reasoning, we introduced explicit step-by-step prompts and 2–4 representative cases as guidance for MLLM.
As shown in Table~\ref{tab:ablation-cot},
CoT yields substantial improvements on the Generation task, as reflected by notable gains in both strong and weaker models (\textbf{GPT-4o}: 30.25\% → 39.36\%; \textbf{Gemma3 (12B)}: 11.75\% → 33.15\%).
However, it brings limited or negative effects on Detection and Understanding, where excessive reasoning often introduces distraction or error propagation.
These results suggest that CoT is most beneficial for tasks that require interpreting user intent and generating multi-step action sequences, but may hinder simpler tasks that depend on direct visual matching due to noisy information in the CoT cases.


\subsubsection{Template-Guided Generation}
In Section~\ref{exp:cot}, we find that Chain-of-Thought enhances the generation ability of MLLMs. To further investigate the factors influencing the quality of PPT generated by VLLM, we introduced an API-level generation template to guide the model's performance in the generation task.

\begin{figure}[h]
	\centering
	\includegraphics[width=1.0\linewidth]{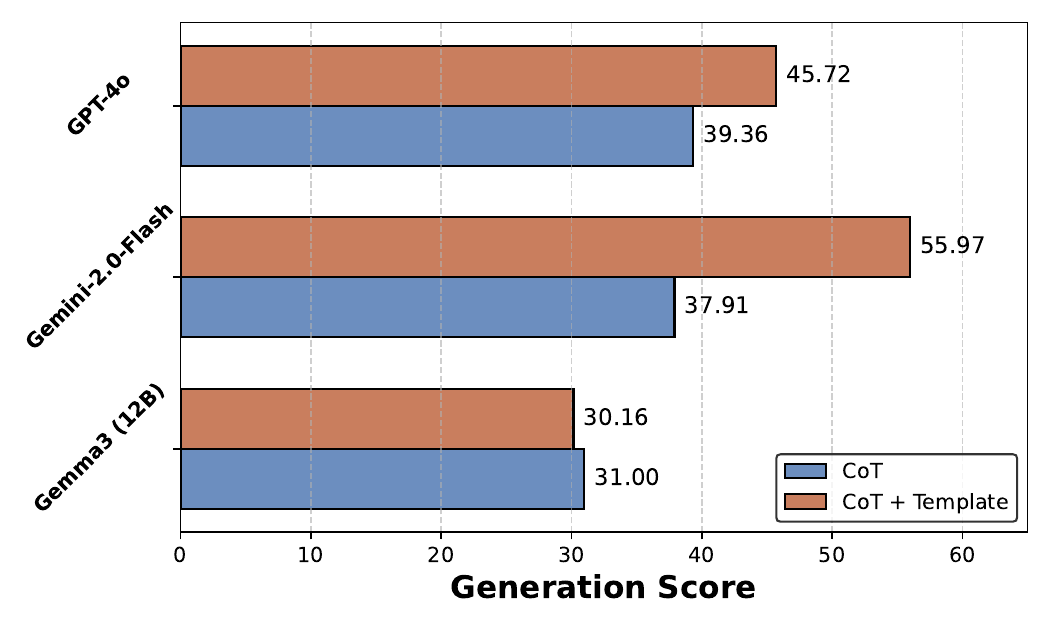}
	\caption{Effect of template-guided prompting on the Generation.}
    \label{fig:generation-template}
\end{figure}

As shown in Figrue~\ref{fig:generation-template}, providing an explicit template substantially improves the quality of the generated slides.
Notably, \textbf{GPT-4o} benefits moderately from templates (39.36\% $\rightarrow$ 45.72\%), and the improvement is even more pronounced for \textbf{Gemini-2.0-Flash}, whose generation accuracy increases from 37.91\% to 55.97\%, demonstrating that even advanced MLLMs gain from external structural templates.
However, for smaller models such as \textbf{Gemma3 (12B)}, template guidance yields inconsistent effects.

Meanwhile, an interesting trend emerges when examining the number of valid API sequences.
When CoT is applied, the count of executable API sequences drops markedly from 748 to 531.
\begin{table}[thbp]
\centering
\caption{LLM vs. Human Scores on Generation Task}
\label{tab:llm-human-generation}
\begin{tabular}{llll}
\hline
\textbf{Judge} & \textbf{LLM} & \textbf{Human} & \textbf{Spearman $\rho$} \\ \hline
OpenAI GPT-4o    & 30.25  & 28.37  & 0.86 \\
Gemini-2.0-Flash & 32.82  & 32.59  & 0.89 \\ \hline
\end{tabular}%
\end{table}
This indicates that while CoT can help MLLMs improve the quality of certain outputs, it also introduces that low-quality or erroneous CoT steps may interfere with the model’s sequential planning ability, leading to execution failures.
We hypothesize that this instability reflects a deeper limitation that current MLLMs still struggle to combine sequential planning with visual understanding, which is essential for generating coherent and well-aligned slides through API-based operations.

\subsubsection{LLM-Judge vs. Human}
Evaluating the Generation task presents inherent challenges due to their multi-dimensional assessment criteria and high degree of subjectivity.
Table~\ref{tab:llm-human-generation} validates the reliability of our LLM-based evaluation by comparing the score of LLM judges and human experts.
The results show that both \textbf{GPT-4o} (30.25\% vs. 28.37\%, $\rho$=0.86) and \textbf{Gemini-2.0-Flash} (32.82\% vs. 32.59\%, $\rho$=0.89) exhibit strong agreement between LLM- and human-judge scores, confirming the reliability of our LLM-based evaluation.

\subsection{Case Study}

A unique aspect of PPTBench is its emphasis on evaluating MLLMs’ ability to reason over slide layout.
To further illustrate current model limitations in this dimension, we show four representative failure cases, each drawn from different task but all centered on layout-related errors.

\begin{figure}[htpt]
	\centering
	\includegraphics[width=0.95\columnwidth]{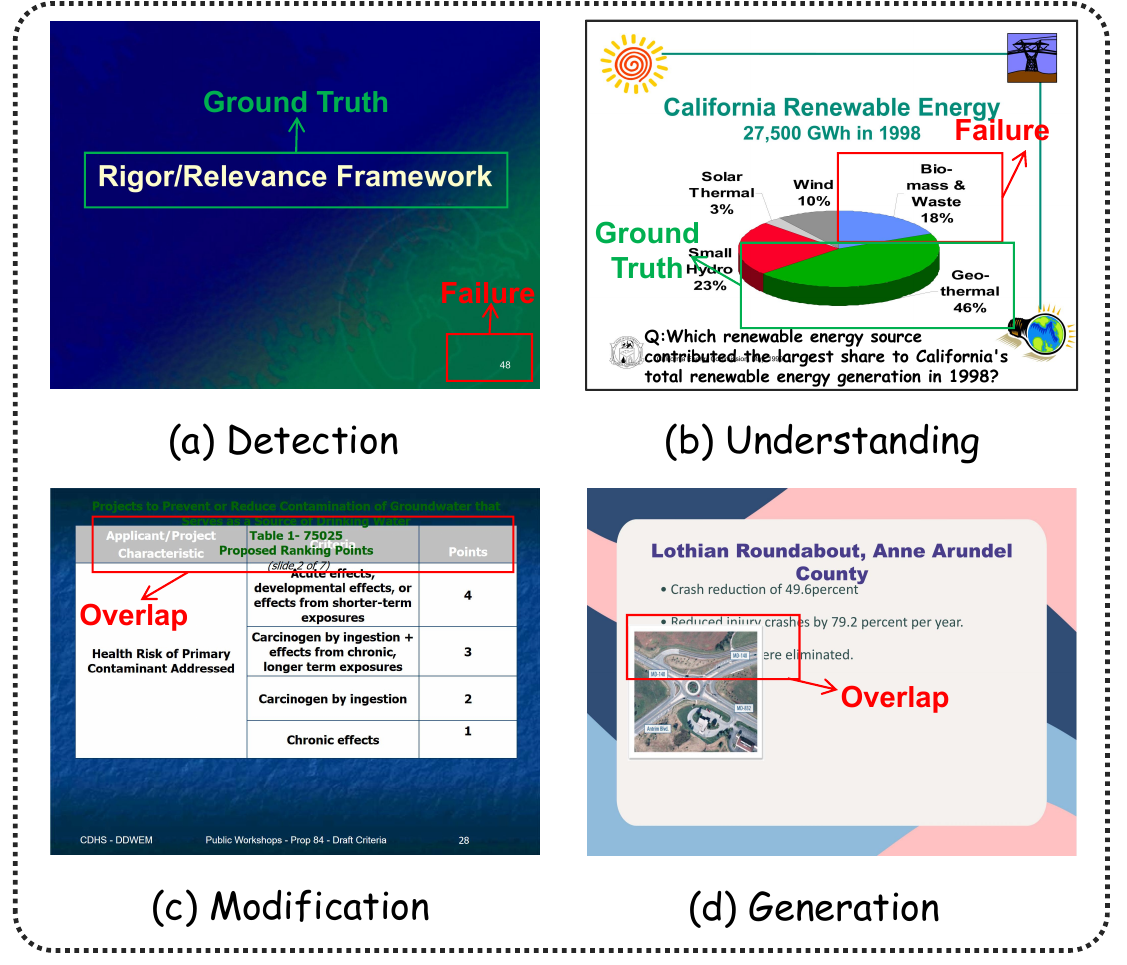} 
	
	\caption{Case study of layout failure cases across four categories. 
	}
	\label{fig:case-studies}
\end{figure}

As shown in Figure~\ref{fig:case-studies},
in \textit{Detection}, the model misidentifies a small slide-number box as the dominant element, overlooking the large central text area.
In \textit{Understanding}, it extracts chart labels but fails to interpret spatial proportions, incorrectly determining the largest sector in a pie chart.
For \textit{Modification}, the model generates an API sequence that moves a table in the wrong direction, causing severe overlap due to inaccurate spatial reasoning.
In \textit{Generation}, the model produces correct textual and visual elements but places the image directly over the bullet points, demonstrating a lack of holistic layout planning.

These examples show that current VLLMs struggle to incorporate layout information into their reasoning processes.
Even when they understand individual elements, they often fail to use spatial structure effectively, limiting their ability to perform PowerPoint-related tasks that depend on accurate geometric and compositional reasoning.

\section{Conclusions}

In this paper, We introduced \textbf{PPTBench}, a comprehensive benchmark that strengthens evaluation on \textit{layout understanding} and \textit{design reasoning}.
It comprises four interrelated tasks—\textit{Detection}, \textit{Understanding}, \textit{Modification}, and \textit{Generation}, enabling a holistic assessment of MLLMs across extracting, reasoning, modification, and creation levels.
These tasks are mutually connected and jointly test models’ abilities at different levels of multimodal comprehension and generation.
Experiments reveal a persistent gap between semantic comprehension and spatial manipulation, reflecting models’ limited ability to coordinate textual semantics with spatial information.
Incorporating \textit{Chain-of-Thought} reasoning and \textit{API templates} brings clear gains in structured editing and layout planning. However, it also indicates that current VLLMs still struggle to integrate sequential planning with visual understanding, limiting their ability to fully exploit API semantics and produce clean, well-organized slides.
In future work, we aim to explore more effective use of templates to achieve tighter fusion among semantics, structure, and visual design for real-world presentation automation.

{
    \small
    \bibliographystyle{ieeenat_fullname}
    \bibliography{main}
}

\clearpage
\setcounter{page}{1}
\maketitlesupplementary

\section{Details on the Original Dataset}
\label{App: datasets}
Our dataset comprises 958 PowerPoint presentations collected from government sources, 
providing a comprehensive foundation for evaluating LLM capabilities. Here, we present 
detailed statistics and analysis of the dataset's composition.

\subsection{Presentation Statistics}
\begin{figure}[t]
    \centering
    \includegraphics[width=\linewidth]{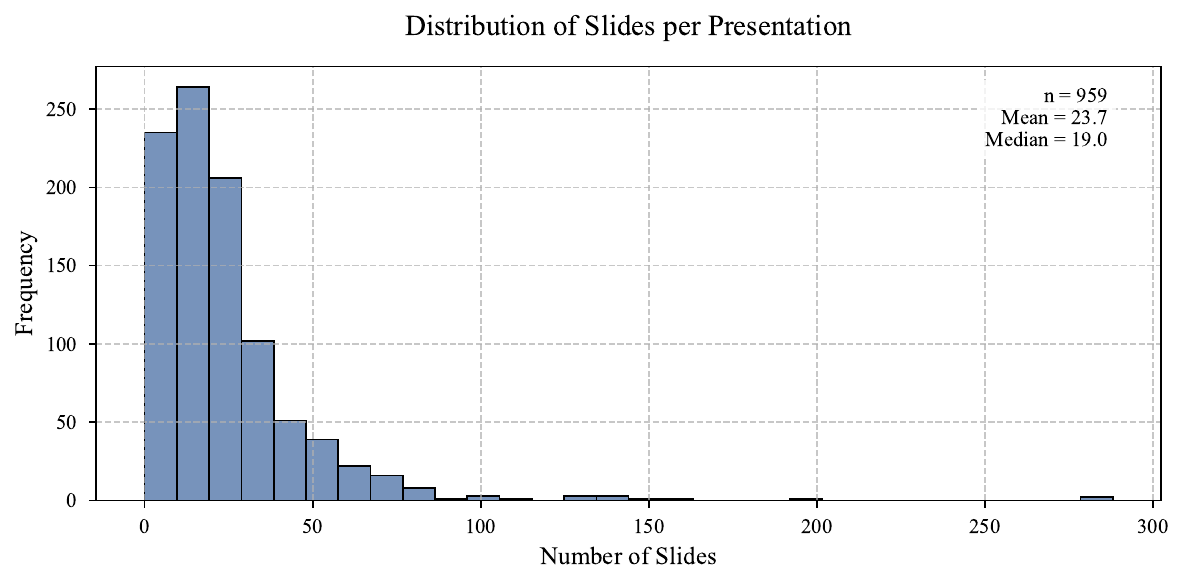}
    \caption{Distribution of slides per presentation. The right-skewed distribution 
    indicates that while most presentations are concise (15--30 slides), there are some 
    longer comprehensive presentations in the dataset.}\label{fig:slides-distribution}
\end{figure}

The dataset shows diverse presentation sizes, with an average of 23.7 slides per 
presentation (median: 19). As shown in Figure~\ref{fig:slides-distribution}, most 
presentations contain 15--30 slides, reflecting typical professional presentation lengths. 
This distribution aligns well with real-world usage patterns, where presentations are 
typically designed for 30--45 minute sessions.

\subsection{Content Element Analysis}
\begin{figure}[t]
    \centering
    \includegraphics[width=\linewidth]{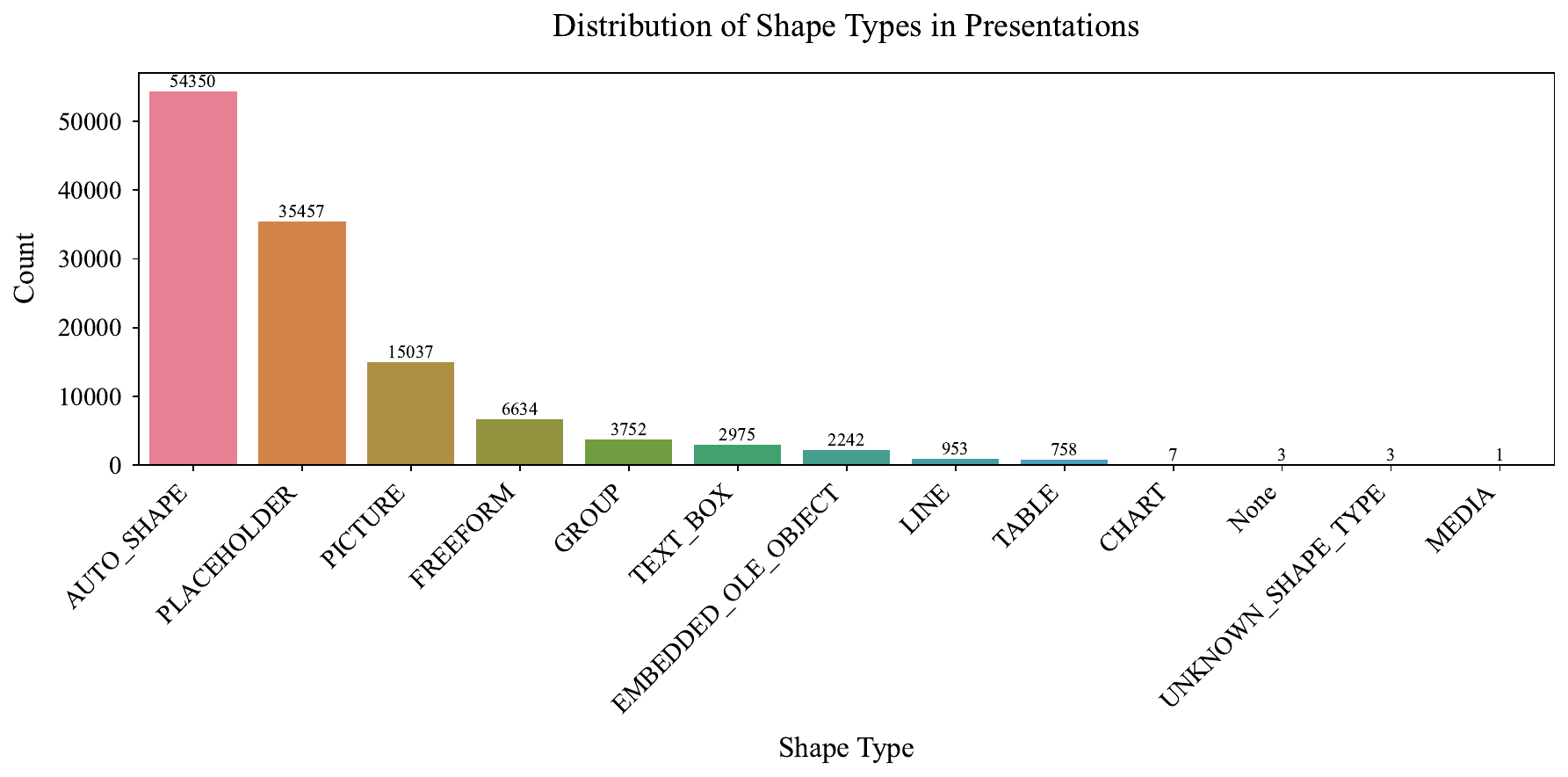}
    \caption{Distribution of shape types across slides, showing the relative frequency 
    of different visual elements used in presentations.}\label{fig:shape-types-distribution}
\end{figure}

The dataset exhibits rich diversity in content elements, providing comprehensive coverage 
for evaluating different aspects of LLM capabilities:

\paragraph{Shape Type Distribution.}
Figure~\ref{fig:shape-types-distribution} reveals the prevalence of different shape types 
across the dataset. Text boxes and basic shapes dominate, accounting for approximately 
60\% of all elements, followed by images (25\%) and charts (15\%). This distribution 
reflects typical presentation composition patterns, where textual content is supported 
by visual elements.

\begin{figure}[t]
    \centering
    \includegraphics[width=\linewidth]{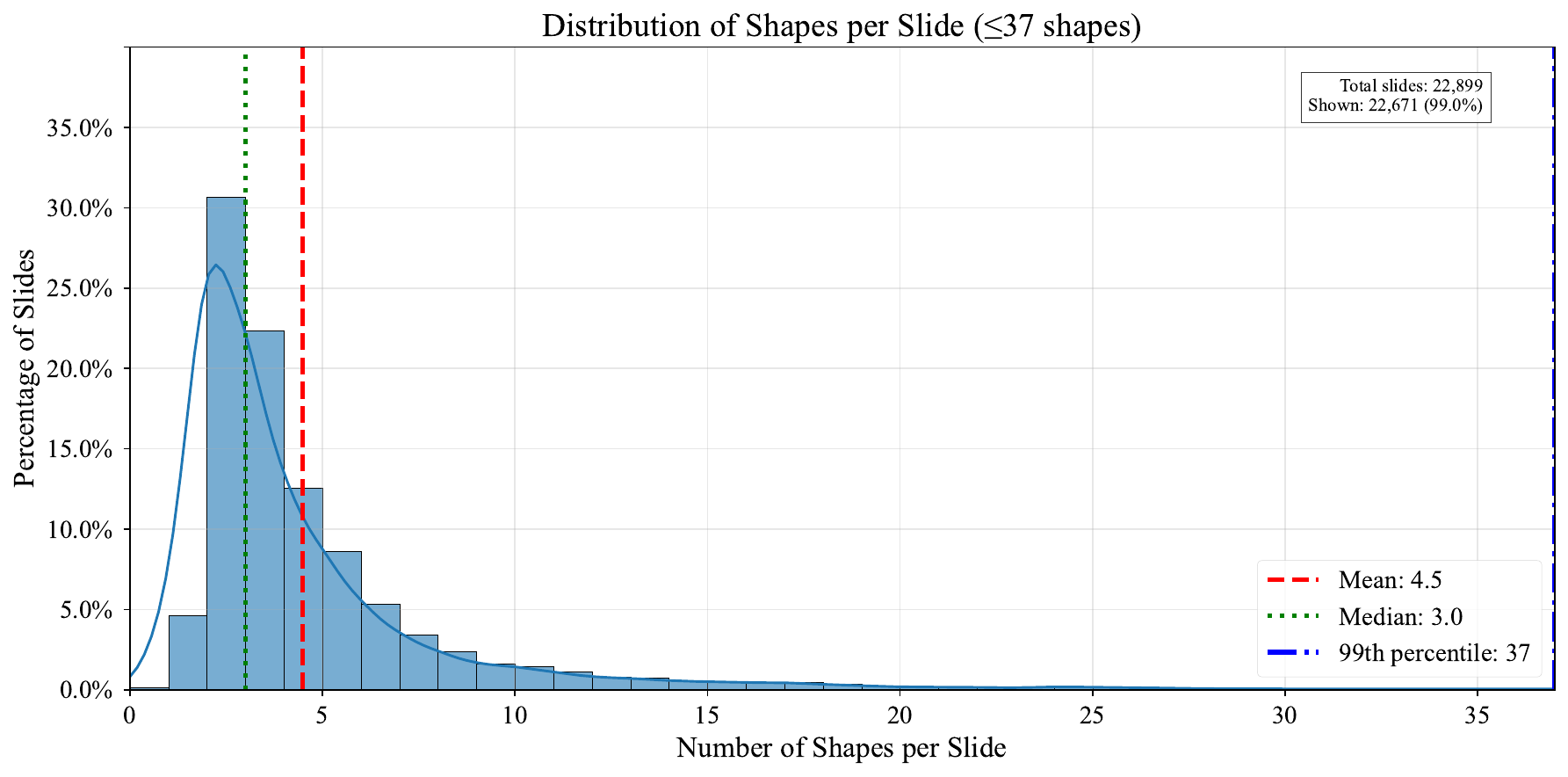}
    \caption{Distribution of shapes per slide, indicating the complexity of slide layouts 
    in the dataset.}\label{fig:shapes-per-slide}
\end{figure}

\paragraph{Slide Complexity.}
The analysis of shapes per slide (Figure~\ref{fig:shapes-per-slide}) indicates moderate 
complexity in slide design. With an average of 4.9 shapes per slide, most slides fall 
within the 5--20 shape range, suggesting well-structured, professional-quality content. 
This complexity level provides sufficient challenge for LLMs while maintaining realistic 
testing scenarios.

\subsection{Dataset Characteristics}
Key characteristics of the dataset include:

\textbf{Domain Coverage:} Predominantly government presentations covering public 
policy, education, and technical topics.

\textbf{Language:} Primary language is English, with consistent professional 
terminology.

\textbf{Temporal Range:} Presentations span from 2015 to 2023, capturing 
evolving presentation styles.

\textbf{Format Consistency:} 4:3 aspect ratio dominates, reflecting institutional 
standardization.

\section{Details on Dataset Construction}
\label{App:data construction}

\subsection{Task Illustrations in PPTBench}
\begin{figure*}[htbp]
    \centering
    \includegraphics[width=0.95\textwidth]{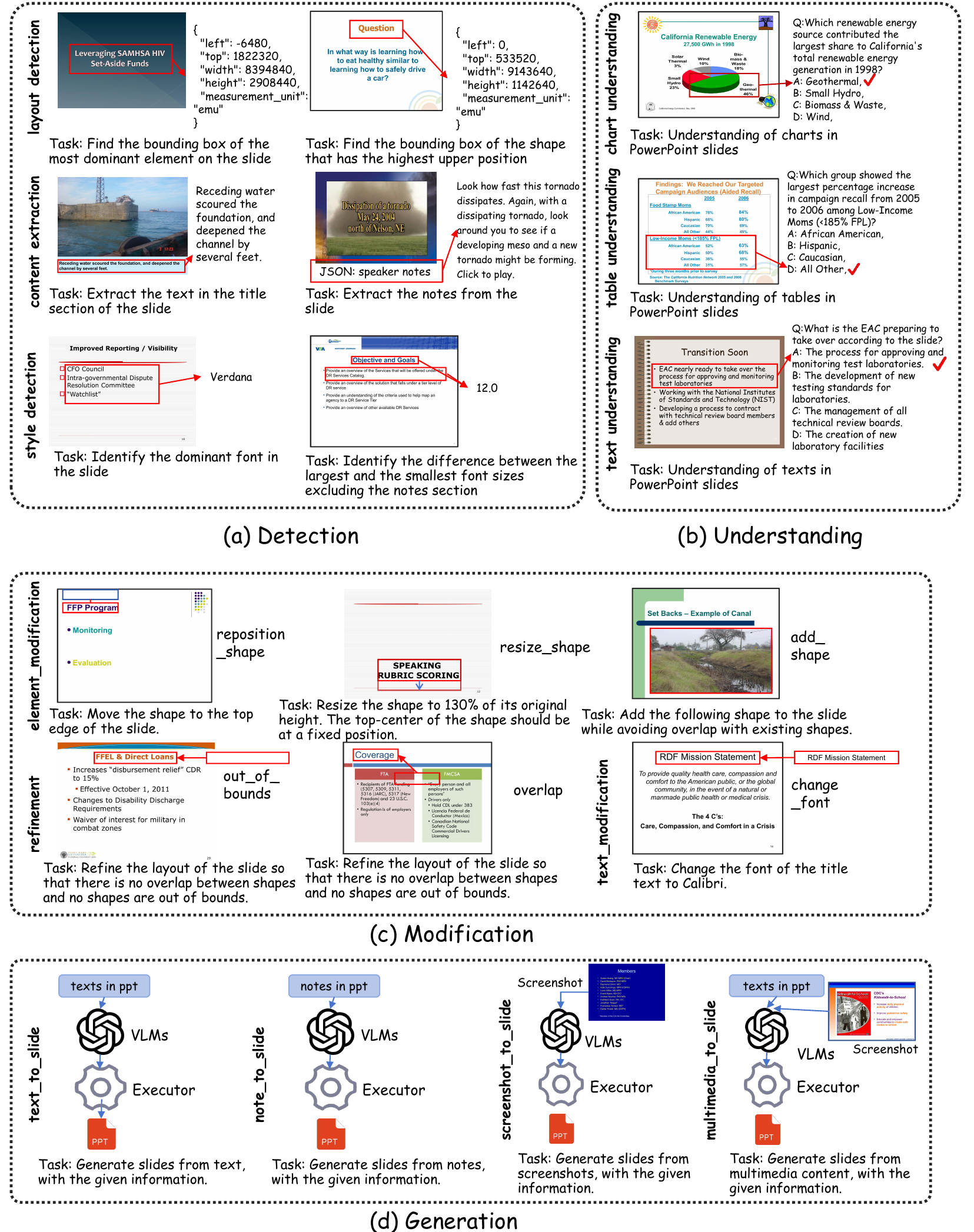}
    \caption{
    Comprehensive illustration of the four major task categories in PPTBench. 
    (a) \textbf{Detection}: element extraction, style detection, and layout detection. 
    (b) \textbf{Understanding}: text, chart, and table understanding and reasoning. 
    (c) \textbf{Modification}: element-level edits, text refinement, and layout adjustment using API operations. 
    (d) \textbf{Generation}: creating slides from text, notes, screenshots, or multimedia inputs via structured API sequences.
    }
    \label{fig:task-overview}
\end{figure*}

To provide a clearer understanding of the four major task categories included in PPTBench, 
we present a comprehensive visual illustration covering representative examples from 
\textit{Detection}, \textit{Understanding}, \textit{Modification}, and \textit{Generation}. 
While Figure~\ref{fig:pptbench-example} in the main paper provides a high-level summary, this appendix includes a 
more detailed and expanded version Figure~\ref{fig:task-overview} to help readers better grasp the diversity and 
realism of the tasks.

\subsection{Structured JSON Representation}
As detailed in the data construction pipeline, we utilize a structured JSON representation to encode both the semantic content and spatial layout of each slide. Listing~\ref{lst:input-json} provides a simplified example of this schema. Note that it explicitly preserves granular attributes such as measurement units (EMUs), precise bounding box coordinates, and hierarchical font details, which are essential for the model to perform accurate layout reasoning and modification.
\begin{lstlisting}[
  caption={Example of the structured JSON input representing a slide.},
  label={lst:input-json},
  basicstyle=\ttfamily\footnotesize]
{
  "slide_width": 9144000,
  "slide_height": 6858000,
  "measurement_unit": "emu",
  "slide": {
    "slide_id": 256,
    "slide_name": "",
    "shapes": [
      {
        "name": "PlaceHolder 1",
        "shape_id": 9,
        "shape_type": "PLACEHOLDER",
        "measurement_unit": "emu",
        "height": 1600200,
        "width": 7772400,
        "left": 762120,
        "top": 2742840,
        "text": "2009 Energy Conference
        April 7, 2009",
        "font_details": [
          {
            "paragraph_index": 0,
            "run_index": 0,
            "text": "2009 Energy Conference",
            "font_name": "Tahoma",
            "font_size": 40
          },
          {
            "paragraph_index": 1,
            "run_index": 0,
            "text": "April 7, 2009",
            "font_name": "Tahoma",
            "font_size": 32
          }
        ],
        "placeholder_type": "SUBTITLE"
      },
      {
        "name": "Picture 4",
        "shape_id": 10,
        "shape_type": "PICTURE",
        "measurement_unit": "emu",
        "height": 1797120,
        "width": 7137360,
        "left": 1143000,
        "top": 380880,
        "auto_shape_type": "RECTANGLE",
        "image_path": "dataset/extracted_images/4B
        YDSTHP4LRBCEE3HACNE3U6APBLJGYE/0
        /image_0_2.png"
      },
      {
        "name": "",
        "shape_id": 11,
        "shape_type": "PICTURE",
        "measurement_unit": "emu",
        "height": 1746360,
        "width": 7013520,
        "left": 1295280,
        "top": 4572000,
        "auto_shape_type": "RECTANGLE",
        "image_path": "dataset/extracted_images/
        4BYDSTHP4LRBCEE3HACNE3U6APBLJGYE/0
        /image_0_3.png"
      }
    ]
  }
}
\end{lstlisting}

\subsection{Construction Details by Task Category}
\paragraph{Detection}
To build a comprehensive detection dataset, we employ different strategies for each task type: 
For content extraction, we systematically extract text elements from PPTX files, preserving 
their hierarchical relationships and structural context. Style detection data is derived 
directly from the embedded metadata in PPTX files, capturing the full range of formatting 
attributes used across presentations.
Layout detection data requires a more sophisticated approach. We first identify slides with 
clean, well-structured layouts as our baseline. Then, using a custom algorithm, we simulate 
common layout issues by introducing controlled modifications:
\textit{Overlapping shapes}, Strategic placement of elements to create intersections; 
\textit{Out-of-bounds elements}, Careful positioning of shapes beyond slide boundaries

\paragraph{Understanding}
For text understanding, we employ human reviewers to select 600 slides with substantial content. 
Using these slides, we generate multiple-choice questions that specifically target the model's 
ability to comprehend textual information and relationships. 
Our construction process involves:
(1) \textbf{Content identification}: Using MLLMs to identify slides containing charts or tables
(2) \textbf{Cross-validation}: Employing multiple MLLMs for accuracy in identification
(3) \textbf{Question generation}: Creating multiple-choice questions using LLMs
(4) \textbf{Answer validation}: Human reviewers verify questions and answers
(5) \textbf{Quality assurance}: Secondary review of 10\% of entries ensures 98\% 
    accuracy rate

\paragraph{Modification}
For element modification tasks, our process involves recording the content and properties of randomly selected shapes, removing them from slides, and creating tasks that require models to reintroduce these elements in appropriate positions.
The text modification dataset is built from slides with substantial textual content. We 
design editing scenarios to evaluate models' capabilities in grammar correction, content 
adaptation, and maintaining consistency with existing slide elements. This approach ensures comprehensive coverage of common text editing requirements in real-world presentations. 
For layout refinement tasks, we identify slides that present opportunities for visual 
enhancement. Our tasks focus on optimizing element alignment and spacing, improving visual hierarchy, and enhancing overall slide balance and composition. These aspects reflect the key considerations in professional presentation design.

\paragraph{Generation}
To build the generation dataset, we extract various content elements from existing presentations:
\textbf{Text content:} Raw text from slides, speaker notes, and annotations; 
\textbf{Visual elements:} Images, charts, diagrams, and other graphical components; 
\textbf{Metadata:} Style information, formatting guidelines, and layout patterns. 
These elements serve as input materials for generation tasks, where models must create new slides that effectively organize and present the provided content while maintaining 
professional design standards. The dataset includes diverse scenarios requiring different types of slide generation, from simple text-based slides to complex multimedia presentations.

\section{Generation Evaluation and Task Examples}

\subsection{Generation Evaluation Examples}
\label{app:generation}
\begin{figure*}[t]
    \centering
    \includegraphics[width=\linewidth]{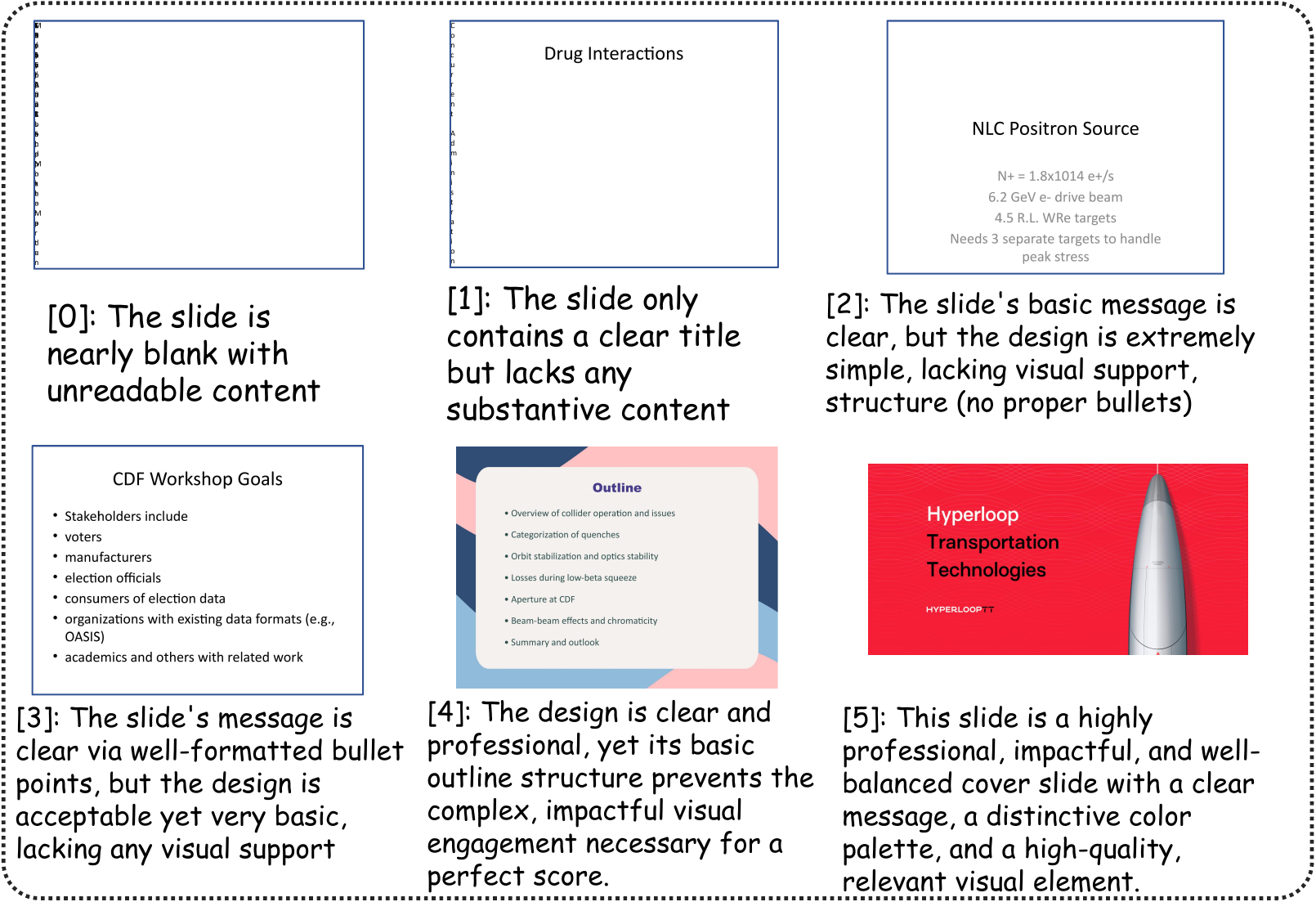}
    \caption{
    Examples of generated slides corresponding to rubric scores 0--5 in the 
    \textit{Generation} task.
    }
    \label{fig:generation-rubric-examples}
\end{figure*}

To better illustrate how the six-point rubric is applied in the \textit{Generation} task, 
we provide representative examples covering all score levels from 0 to 5. 
As shown in Figure~\ref{fig:generation-rubric-examples}, these examples highlight how differences in content clarity, layout organization, and visual design influence the final ratings assigned by the LLM judge. 
This visualization also helps clarify the qualitative distinctions between adjacent score 
levels—particularly between mid-range (2–3) and high-quality (4–5) outputs.

\subsection{Representative Task Instances}
\label{app:task-instances}

\begin{figure*}[t]
    \centering
    \includegraphics[width=\linewidth]{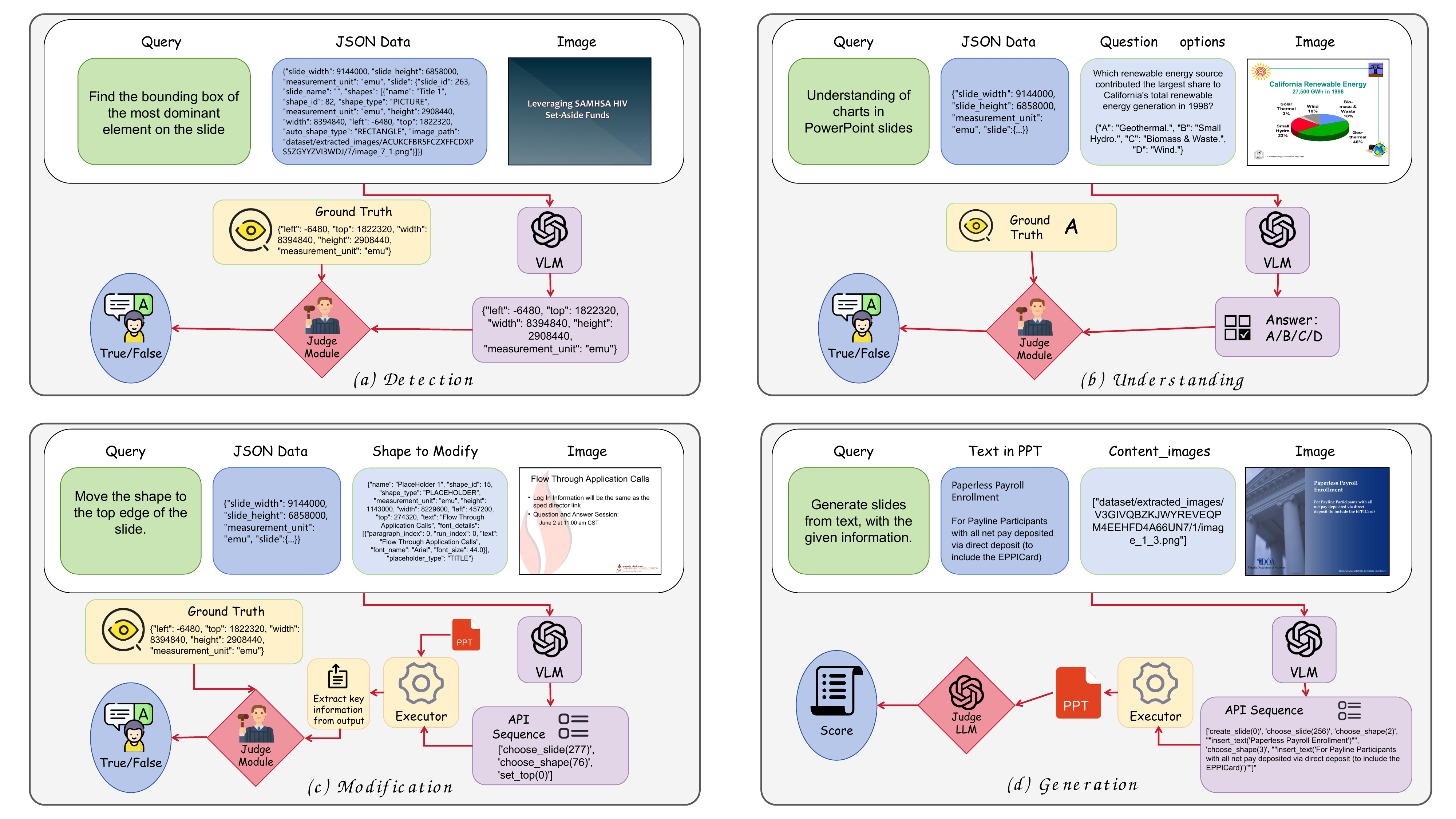}
    \caption{
    Representative task instances across the four PPTBench categories.
    }
    \label{fig:pptbench-task-instances}
\end{figure*}

To provide a clearer view of how each task operates within PPTBench, 
Figure~\ref{fig:pptbench-task-instances} illustrates representative examples 
from all four task categories. Each instance includes the slide image, 
its corresponding structured JSON data, the task-specific query, the model's response format, and the ground-truth label or API sequence.

\section{Details on Experiments}

\subsection{Computational Resource Details}
For all experiments, we utilized a consistent hardware environment consisting of eight NVIDIA Tesla A100 GPUs (40GB VRAM each) and two Intel Xeon 12-core CPUs operating at 3.0GHz with 256GB RAM.\@
The system ran Ubuntu 20.04.5 LTS with CUDA 12.4. 

For closed-source models, we accessed APIs with consistent parameters (temperature=0.0, top\_p=0.95) to ensure reproducibility. 
For open-source models (Llama3.2-Vision, LLaVA-1.5, and MiniCPM-V), we deployed them using the ollama framework\footnote{https://ollama.com} with 4-bit quantization to optimize memory usage while maintaining inference quality. 
All inference runs were configured with a consistent maximum context length of 2,048 tokens and batch size of 1 to ensure uniform evaluation conditions across all models.

\subsection{Prompt Templates and API Specifications}
\label{App:Templates and API}

To ensure consistent evaluation across all PPTBench tasks, we design standardized 
prompt templates and a unified API schema that define how models interpret task 
instructions and execute editing or generation operations in a deterministic manner.

\paragraph{Prompt Templates for All Tasks.}
We developed standardized prompt templates across all task categories. 
Each template provides clear task definitions, representative examples, and explicit 
output format constraints to reduce linguistic ambiguity.  
Listing~\ref{lst:detection-prompt-template} illustrates the prompt design for style detection, 
while Listing~\ref{lst:modification-prompt-template} demonstrates the constrained API-based format used in modification tasks.

\paragraph{API Schema and Design Principles.}
PPTBench provides a standardized set of 17 atomic PowerPoint API functions 
(Listing~\ref{lst:api-calls}) that allow models to manipulate slide content in a fully executable and deterministic manner. 
These APIs follow three essential principles:  
\textbf{Atomicity}—each operation performs a single unambiguous action;  
\textbf{Composability}—complex edits or generation procedures can be achieved by 
combining atomic operations;  
\textbf{Executability}—all functions map directly to deterministic slide-level updates 
via our PowerPoint-compatible executor.  
This standardization ensures that model performance differences stem from capability 
differences rather than prompt or implementation variance.

\paragraph{API Templates for Generation.}
To further analyze structure-aware Generation, we introduce a set of API templates that 
encode canonical layout patterns and serve as structural priors for template-guided 
generation.  
Each template is composed of deterministic API sequences specifying slide layout, 
background, typography, and element positioning.  
We include three representative templates: a \textit{Title Slide} template that establishes 
a clear visual hierarchy (Listing~\ref{lst:generation-api-title}), a 
\textit{Bullet Points Slide} template focusing on textual alignment and spacing 
(Listing~\ref{lst:generation-api-bullet}), and an \textit{Image with Description Slide} 
template illustrating visual–textual coordination 
(Listing~\ref{lst:generation-api-image}).  
These templates are used in the template-guided generation ablation 
(see Figure~\ref{fig:generation-template}), enabling controlled comparison between 
free-form and structurally guided generation.

\begin{lstlisting}[
  breaklines=true,
  caption={API-level template for a \textit{Title Slide}.},
  label={lst:generation-api-title},
  columns=fullflexible,
  basicstyle=\ttfamily\small]
Title Slide:
[
"create_slide(6)",
"add_picture(0, 0, 720, 540, 'dataset/A modern, geometric design featuring a central rounded frame bordered by blocks of navy, pastel pink, and light blue.jpg')", # Background Image
"add_text_box(60, 160, 600, 100, 'Science, Technology, Engineering\\nand Mathematics Talent\\nExpansion Program (STEP)')", # Our Main Title
"set_font_color('483D8B')",
"set_font_size(32)",
'set_text_align("CENTER")',
"set_font(font_name='Arial Black')",
"add_text_box(110, 280, 500, 50, 'National Science Foundation Directorate for Education & Human Resources Division of Undergraduate Education (DUE)')", #our Subtitle Description
"set_font_color('2F4F4F')",
"set_word_wrap(True)",
"set_font_size(24)",
"set_font(font_name='Calibri')",
"add_text_box(210, 415, 300, 30, 'November 18, 2009')",
"set_text_align('CENTER')",
"set_font_color('696969')",
"set_font_size(18)",
"set_font(font_name='Calibri')"
]
  
\end{lstlisting}

\begin{lstlisting}[
  breaklines=true,
  caption={API-level template for a \textit{Bullet Points Slide}.},
  label={lst:generation-api-bullet},
  columns=fullflexible,
  basicstyle=\ttfamily\small]
Bullet_points Template:
[
"create_slide(6)",
"add_picture(0, 0, 720, 540, 'dataset/A modern, geometric design featuring a central rounded frame bordered by blocks of navy, pastel pink, and light blue.jpg')",
"add_text_box(80, 70, 560, 60, 'Some Guiding Principles (Water 2025)')",
"set_text_align('CENTER')",
"set_font_color('483D8B')",  # Same color as the main title for visual emphasis
"set_font_size(28)",
"set_font(font_name='Arial Black')",
"add_text_box(100, 130, 520, 360, "
"'Recognize and respect state, tribal, and federal water rights, contracts, and interstate compacts or decrees of the United States Supreme Court that allocate the right to use water.\\n\\n' "
"'Maintain and modernize existing water facilities so they will continue to provide water and power.\\n\\n' "
"'Enhance water conservation, use efficiency, and resource monitoring to allow existing water supplies to be used more effectively.\\n\\n')",
"set_text_align('LEFT')",
"set_word_wrap(True)",
"set_font_color('2F4F4F')", 
"set_font_size(22)",
"set_font(font_name='Calibri')"
]
\end{lstlisting}

\begin{lstlisting}[
  breaklines=true,
  caption={API-level template for an \textit{Image with Description Slide}.},
  label={lst:generation-api-image},
  columns=fullflexible,
  basicstyle=\ttfamily\small]
Image_With_Description Template:
[
    "create_slide(6)",
    "add_picture(0, 0, 720, 540, 'dataset/A modern, geometric design featuring a central rounded frame bordered by blocks of navy, pastel pink, and light blue.jpg')",
    "add_picture(60, 70, 280, 400, 'dataset/extracted_images/5GLBKNRC3EXYC6UNLASJOTJJ7XBA7MA6/2/image_2_1.jpg')", 
    "add_text_box(380, 90, 280, 80, '"Tosoiba" - Land of sparkling waters')",
    "set_word_wrap(True)",
    "set_font_color('483D8B')",
    "set_font_size(32)",
    "set_font(font_name='Arial Black')",
    "add_text_box(380, 240, 280, 150, 'Understanding the hydro-geology in and around Monsanto's operations has been a key component of our business for decades now.')",
    "set_word_wrap(True)",
    "set_font_color('2F4F4F')",  # Secondary color for body text
    "set_font_size(18)",
    "set_font(font_name='Calibri')",]
\end{lstlisting}

\subsection{Chain-of-Thought Examples}
\label{app:cot-examples}

While CoT prompting is applied across all four task categories in PPTBench, 
its impact is most pronounced in tasks that require multi-step planning and spatial reasoning. 
Accordingly, we focus here on representative CoT examples for the 
\textit{Modification} and \textit{Generation} tasks, where CoT directly guides the model through geometric reasoning, layout refinement, and sequential API execution.

\begin{lstlisting}[
  caption={Chain-of-Thought example for font modification in the \textit{Modification} task.},
  label={lst:cot-font-modification},
  basicstyle=\ttfamily\footnotesize,
  breaklines=true
]
{
    "query":"Change the font of the title text to Arial Black.",
    
    "reasoning": "Locate the slide and the title shape: the slide object has slide_id 311 and a shape with placeholder_type 'TITLE' and shape_id 722. The title's current font is 'Tahoma' at 24.0pt, so to change the title font to 'Arial Black' we should (1) select the slide, (2) select the title shape, and (3) set the font to 'Arial Black'. No size or position changes are required.",
    "functions": [
        "choose_slide(311)",
        "choose_shape(722)",
        "set_font('Arial Black')"
}
\end{lstlisting}

\begin{lstlisting}[
  caption={Chain-of-Thought example for slide creation in the \textit{Generation} task.},
  label={lst:cot-generation-example},
  basicstyle=\ttfamily\footnotesize,
  breaklines=true
]
{
    "query": "Generate slides from multimedia content, with the given information.", 
    
    "reasoning": "Goal: produce two slides from the multimedia materials - a title slide and a content slide with a large image. Steps: (1) Create a title slide and select the appropriate layout: call `create_slide(0)` to make a new slide and `choose_slide(256)` to set the Title Slide template. (2) Populate the title placeholder: select the title shape (`choose_shape(2)`) and insert the title text 'Under Dispenser Containment' so the first slide clearly states the topic. (3) Create a second slide for the image: call `create_slide(8)` to add a new slide using an image-centric layout, then `choose_slide(257)` to pick the specific template instance. (4) Add a consistent page title on the second slide: select its title placeholder (`choose_shape(2)`) and insert the same title to maintain context across slides. (5) Insert the main image into the image placeholder: switch to the image placeholder (`choose_shape(3)`) on slide 257 and call `add_picture(...)` with coordinates and size that place the image centrally and make it the dominant visual element while preserving margins. (6) Keep the reasoning concise and output the ordered function sequence afterwards.",
    "functions": ['create_slide(0)',
        'choose_slide(256)',
        'choose_shape(2)',
        "insert_text('Under Dispenser Containment')",
        'create_slide(8)',
        'choose_slide(257)',
        'choose_shape(2)',
        "insert_text('Under Dispenser Containment')",
        'choose_slide(257)',
        'choose_shape(3)',
        "add_picture(141, 48, 432, 324, 'dataset/extracted_images/W7RJCH3WN2DEJH5CHAXOMF7RX3AFE3PW/17/image_17_2.jpg')"]
}
\end{lstlisting}

\begin{lstlisting}[
  float=*,
  breaklines=true,
  caption={Prompt template for the detection category, style detection subcategory.},
  label={lst:detection-prompt-template},
  columns=fullflexible,
  basicstyle=\ttfamily\small]
prompt = f"""
Task: You are given a slide from a presentation in the form of an image and JSON data.\n
{_QUERY_}. Provide only the requested information in JSON format without any additional text or explanations.\n
Examples:
Query: Identify the dominant font in the slide.
Answer:\n{_EXAMPLE1_}\n
Query: Identify the difference of the largest and the smallest font size excluding the notes section.
Answer:\n{_EXAMPLE2_}\n
{_DIVIDER_}\n
Slide JSON: {slide_json}\n
{_DIVIDER_}\n
Query: {_QUERY_}
Answer (Please provide a JSON object with the key "answer" and the value being the answer to the query, without any additional text):
"""
\end{lstlisting}

\begin{lstlisting}[
  float=*,
  breaklines=true,
  caption={Prompt template for the modification category, refinement subcategory.},
  label={lst:modification-prompt-template},
  columns=fullflexible,
  basicstyle=\ttfamily\small]
prompt = f"""
Task: You are given a slide from a presentation in the form of an image and JSON data.\n
{_QUERY_}. 
IMPORTANT: You are ONLY allowed to use the following functions. Any other functions or operations are not allowed:
{_API_LIST_}
These are the only valid functions you can use. Do not attempt to use any other functions or operations.\n
Required format:
- Return ONLY a valid JSON dictionary
- No explanation text before or after the JSON
- No markdown formatting\n
Examples:
Examples:
{
  "function1": "choose_slide(0)",
  "function2": "choose_shape(1)",
  "function3": "set_width(1000000)",
  "function4": "insert_text('Hello, World!')"
}\n
{_DIVIDER_}\n
Slide JSON: {slide_json}\n
{_DIVIDER_}\n
Query: {_QUERY_}
Answer:
"""
\end{lstlisting}

\begin{lstlisting}[
  float=*,
  breaklines=true,
  caption={Complete set of API calls given to the model along with the descriptions},
  label={lst:api-calls},
  columns=fullflexible,
  basicstyle=\ttfamily\small]
api_list = """
create_slide() Description: Create a new slide. Notes: This function creates a new slide in the presentation.
choose_shape(shape_id: int) Description: Choose a shape to work with. Notes: The shape_id value can be found with the key 'shape_id' in the JSON data. The slide must be selected before calling this function.
choose_slide(slide_id: int) Description: Choose a slide to work with. Notes: The slide_id value can be found with the key 'slide_id' in the JSON data.
set_width(width: int) Description: Set the width of the selected shape. Notes: The shape must be selected before calling this function.
set_height(height: int) Description: Set the height of the selected shape. Notes: The shape must be selected before calling this function.
set_top(top: int) Description: Set the top of the selected shape. Notes: The shape must be selected before calling this function.
set_left(left: int) Description: Set the left of the selected shape. Notes: The shape must be selected before calling this function.
add_text_box(left: int, top: int, width: int, height: int, text: str) Description: Add a text box to the current slide, and choose the shape as the current shape. Notes: The current slide must be set before calling this function.
add_picture(left: int, top: int, width: int, height: int, image_path: str) Description: Add a picture to the current slide, and choose the shape as the current shape. Notes: The current slide must be set before calling this function.
insert_text(text: str) Description: Insert text into the selected shape. Notes: The shape must be selected before calling this function.
set_font_size(font_size: int) Description: Set the font size of the selected shape. Notes: The shape must be selected before calling this function.
set_font_style(font_style: str) Description: Set the font style of the selected shape. Notes: The shape must be selected before calling this function.
set_font(font_name: str) Description: Set the font of the selected shape. Notes: The shape must be selected before calling this function.
set_font_color(font_color: str) Description: Set the font color of the selected shape. Notes: The shape must be selected before calling this function.
set_notes_font(font_name: str) Description: Set the font of the notes in the presentation. Notes: The slide must be selected before calling this function, and it does not require a shape to be selected.
set_text_align(align_style: str) Description: Set the horizontal alignment of the text in the selected shape. Notes: The shape must be selected *before* calling this function (using `choose_shape`). This alignment is applied to all paragraphs within the currently selected shape, and *only* affects that shape.
set_word_wrap(wrap: bool = True) Description: Enables or disables automatic word wrapping for the text within the currently selected shape. Notes: A shape that can contain text (like a textbox or an autoshape) must be selected before calling this function.
"""
\end{lstlisting}

\end{document}